\documentclass{article}

\usepackage[preprint,nonatbib]{neurips_2020}

\usepackage[utf8]{inputenc} % allow utf-8 input
\usepackage[T1]{fontenc}    % use 8-bit T1 fonts
\usepackage{hyperref}       % hyperlinks
\usepackage{url}            % simple URL typesetting
\usepackage{booktabs}       % professional-quality tables
\usepackage{amsfonts}       % blackboard math symbols
\usepackage{nicefrac}       % compact symbols for 1/2, etc.
\usepackage{microtype}      % microtypography
\usepackage{xcolor}
\usepackage{bm}
\usepackage{bbm}
\usepackage{amsbsy}
\usepackage{amsmath}
\usepackage{enumitem}
\usepackage{lipsum}
\usepackage{caption}
\usepackage{graphicx}
\usepackage{subcaption}
\usepackage[titletoc]{appendix}
\usepackage{grffile}
\usepackage{wrapfig}

\DeclareMathAlphabet{\altmathcal}{OMS}{cmsy}{m}{n}

\usepackage{color,colortbl}
\definecolor{lightgray}{HTML}{F0F0F0}  %F3F3F3
\definecolor{rowbackground}{HTML}{F9F9F9}
\definecolor{Gray}{gray}{0.95}
\definecolor{LightCyan}{rgb}{0.88,1,1}

\newtheorem{property}{Property}
\newcolumntype{a}{>{\columncolor{Gray}}l}

\usepackage{multirow}
\usepackage{adjustbox}

\usepackage{xspace}
\newcommand{\method}{\texttt{ELF}\xspace}

\definecolor{electricultramarine}{rgb}{0.25, 0.0, 1.0}
\definecolor{forestgreen}{rgb}{0.13, 0.55, 0.13}
\definecolor{darkgreen}{rgb}{0.0, 0.5, 0.0}
\definecolor{brightgreen}{rgb}{0.4, 1.0, 0.0}

\title{ELF: An Early-Exiting Framework\\ for Long-Tailed Classification}

\author{%
  Rahul Duggal \\
  Georgia Tech \\
  \texttt{rahulduggal@gatech.edu} \\
  \And
  Scott Freitas \\
  Georgia Tech \\
  \texttt{safreita1@gatech.edu} \\
  \AND
  Sunny Dhamnani \\
  Georgia Tech \\
  \texttt{sdhamnani3@gatech.edu} \\
  \And
  Duen Horng (Polo) Chau \\
  Georgia Tech \\
   \texttt{polo@gatech.edu} \\
  \And
  Jimeng Sun \\
  UIUC \\
   \texttt{jimeng@illinois.edu} \\
}

\begin{document}

\maketitle

\begin{abstract}

The natural world often follows a long-tailed data distribution where only a few classes account for most of the examples. 
This long-tail causes classifiers to overfit to the majority class.
To mitigate this, prior solutions commonly adopt class rebalancing strategies such as data resampling and loss reshaping. 
However, by treating each example within a class equally, these methods fail to account for the important notion of \textit{example hardness}, i.e., within each class some examples are easier to classify than others.
To incorporate this notion 
of hardness
into the learning process, we propose the \textit{EarLy-exiting Framework} (\method{}). 
During training, \method{} learns to early-exit easy examples through auxiliary branches attached to a backbone network. 
This offers a dual benefit---(1) the neural network increasingly focuses on hard examples, since they contribute more to the overall network loss; and (2) it frees up additional model capacity to distinguish difficult examples. 
Experimental results on two large-scale datasets, ImageNet LT and iNaturalist'18, demonstrate that \method{} can improve state-of-the-art accuracy by more than 3\%.
This comes with the additional benefit of reducing up to 20\% of inference time FLOPS. 
\method{} is complementary to prior work and can naturally integrate with a variety of existing methods to tackle the challenge of long-tailed distributions. 
% We provide our code as part of the supplementary material.

\end{abstract}

\section{Introduction}
Real data often follows a long-tailed distribution where the majority of examples are from only a few classes.
On datasets following this distribution, neural networks often favor the majority class, leading to poor generalization performance on rare classes.
This imbalance problem has traditionally been solved by resampling the data (undersampling, oversampling) \cite{zhou2019bbn,kang2019decoupling,peng2019trainable,chawla2002smote,he2008adasyn}, or reshaping the loss function (loss reweighting, regularization) \cite{cui2019class,cao2019learning}. 
However, these existing approaches focus on \textit{class size} to address the challenge of data imbalance, without taking into account the ``\textbf{hardness}'' of each example within a class.
Intuitively, there might be easy examples in the minority classes that get incorrectly up-weighted, or difficult examples in the majority classes that get erroneously down-weighted. 
We show that by incorporating this notion of example hardness during training our method can (correctly) increase the loss contribution of hard examples across all classes.

\begin{figure}[t!]
\centering
    % \begin{minipage}[t]{0.47\textwidth}
    % \includegraphics[width=0.65\linewidth]{NEURIPS 2020/Images/acc_vs_FLOPS2.pdf}
    \includegraphics[width=\textwidth]{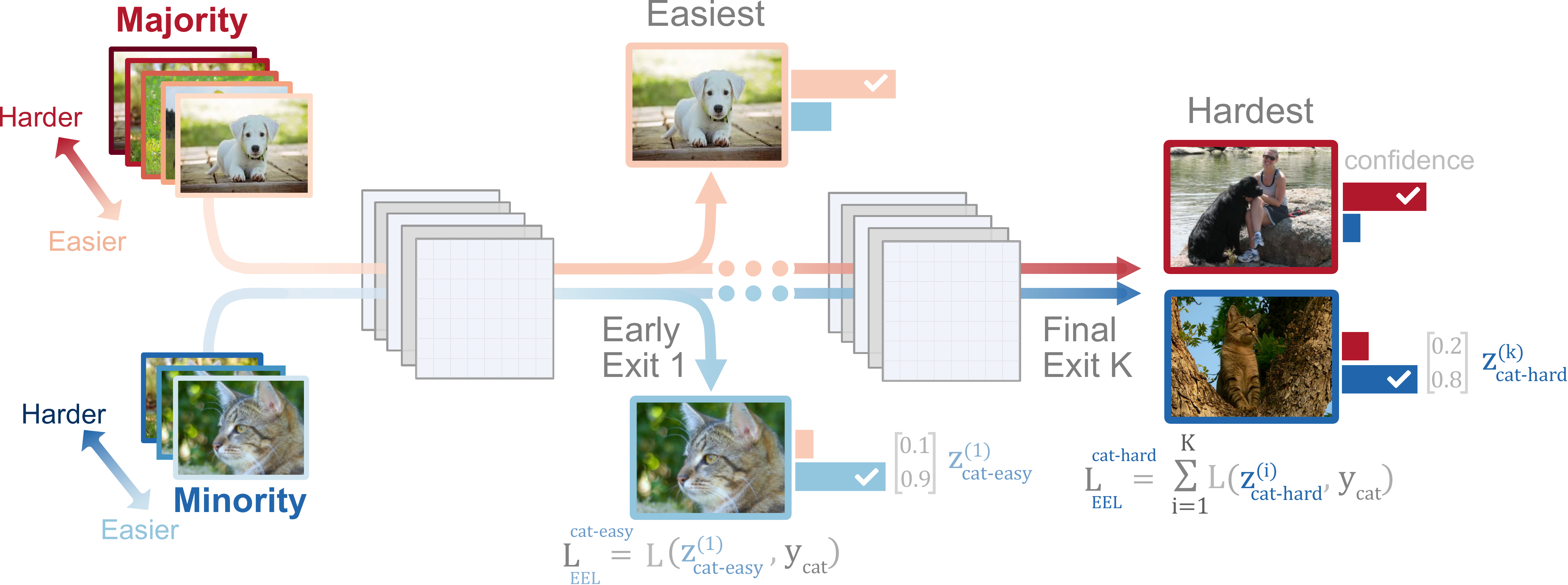}
   \caption{Our \textit{EarLy-exiting Framework} (\method) augments a backbone network with auxiliary classifier branches. 
    During training, \method{} aims to confidently and correctly classify examples at the earliest possible exit branch.
    Harder examples exit later in the network, accumulating a higher overall loss. 
    % \method{} enables on-the-fly model selection based on available compute budget.
    }
    \label{fig:crown}
\end{figure}

We propose the \textit{EarLy-exiting Framework} (\method{}) (Figure~\ref{fig:crown}) to incorporate this notion of example hardness during training.
\method{} is premised around the idea of \textit{learning} to exit ``easy'' examples earlier in the network and ``harder'' examples towards the end.
To achieve this, \method attaches auxiliary classifier branches to a backbone network which we refer to as early-exits.
% \from{srijan}{all}{define auxiliary exit first before using it.}
At each early-exit, the neural network tries to correctly predict the input with high confidence. 
If the prediction is incorrect or the confidence is not high enough (typical for difficult inputs), the example incurs a loss and proceeds to the next exit. 
Our proposed \textbf{early-exiting during training} has several advantages:

\begin{enumerate}[leftmargin=*]
    \item \textbf{Shifting model focus} towards harder examples by increasing the average loss contribution of difficult examples compared to easier ones.
    
    \item \textbf{Freeing model capacity} to focus on harder examples by exiting easier ones early in the network.
    
    \item \textbf{Computational savings} during inference by reducing the average FLOPS required per image.
    
    \item \textbf{Enabling on-the-fly} model selection for variable compute budget.
\end{enumerate}

The concept of early-exiting has traditionally been used during inference to reduce the number of floating-point operations (FLOPS) and save energy~\cite{teerapittayanon2016branchynet,huang2017multi}.
In contrast, \method{} uses early exiting during \textit{training} to learn the concept of example hardness which helps with the problem of class imbalance.

\noindent
\textbf{Contributions.} Our contributions are four-fold:
(1) we identify the key concept of example hardness to help improve generalization performance under long-tailed data distributions;
(2) we propose \method{}, a generic framework that complements existing research by incorporating the notion of example hardness during training;
(3) we demonstrate that \method{} can generate a family of models to enable on-the-fly model selection for variable computate budgets;
(4) we perform extensive evaluation on large-scale imbalanced datasets ImageNet LT and iNaturalist'18, improving the state-of-the-art imbalanced classification accuracy by more than 3\%, and reducing FLOPS by up to 20\%.

\section{Related Research}
Recent research has increasingly shifted focus from classification on artificially balanced datasets \cite{deng2009imagenet, duggal2019cup} to classification under a long-tail class distribution \cite{liu2019large, duggal2020rest}. We discuss closely related research from the areas of long-tailed classification and early-exiting.

\textbf{Long-tailed classification.}
 Prior work in this area can be categorized along the following three direction---(1) data resampling, (2) loss reshaping, and (3) transfer and meta learning.

Data resampling approaches class imbalance by either repeating examples for the rare class (oversampling) \cite{wilson1972asymptotic, chawla2002smote, han2005borderline, he2008adasyn} or discarding existing examples from the majority class (undersampling)~\cite{wilson1972asymptotic, yen2009cluster}. 
The distinguishing factor among these methods lies in the criteria used for under(over)sampling. 
For example, SMOTE~\cite{chawla2002smote} oversamples the minority class through linear interpolation, whereas \cite{yen2009cluster} undersamples by clustering the majority class examples and replacing them with a few anchor points. 
However, oversampling generally creates redundancy and 
risks overfitting to rare classes.
On the other hand, undersampling is susceptible to losing information from the majority classes~\cite{buda2018systematic}.

Loss reshaping tackles class imbalance by either \textit{reweighting} example loss, or through class dependent \textit{regularization}. 
Reweighting based approaches assign a larger weight to rare class examples and a smaller weight to the majority ones~\cite{cui2019class, huang2019deep, wang2017learning, zhang2017range, dong2017class, huang2016learning}. 
On the other hand, regularization methods tackle class imbalance by establishing margins depending upon the class size~\cite{liu2016large, liu2017sphereface, wang2018additive}.
Recently, \cite{cao2019learning} proposed a loss reshaping method (LDAM) to learn larger margins around rare classes.
Among loss reshaping methods, the closest to \method{} is the Focal loss~\cite{lin2017focal}, which incorporates a notion of input-hardness by reweighting an example's loss in proportion to its prediction confidence.
However, in practice Focal loss does not generalize well in the long-tailed setting.
\method{} complements these existing loss reshaping methods, further improving accuracy under class imbalance.

Transfer and meta learning based approaches aims to transfer knowledge from the majority classes to rare ones by transfer learning, multi task learning or learning to learn~\cite{liu2019large, yin2019feature, cui2018large}.
However, it is observed that these approaches are generally more computationally expensive than loss reshaping methods~\cite{cao2019learning}, including our proposed method \method{}.

\textbf{Early exiting in neural networks.}
Research in this area focuses on reducing computation during inference by dynamically routing examples through a network based on their hardness 
% \scott{use this wording everywhere, much stronger}
~\cite{teerapittayanon2016branchynet,Huang2018MultiScaleDN,phuong2019distillation,hu2020triple}. 
The core intuition behind these methods is to reduce the computation during \textit{inference} by routing easier examples through early exits.
Different papers define the notion of hardness differently.
For example, \cite{teerapittayanon2016branchynet} defines hardness using the entropy of the prediction vector so that easier examples with low entropy predictions exit earlier. 
On the other hand \cite{Huang2018MultiScaleDN,phuong2019distillation,hu2020triple} defines hardness based on prediction confidence. 
Thus, (easier) examples with high prediction confidence exit earlier.
% These methods differ in the criteria they use to establish hardness during inference.
% For example \cite{teerapittayanon2016branchynet} define hardness based on entropy of the predicted probabilities so that easy examples have low entropy. \cite{Huang2018MultiScaleDN,phuong2019distillation,hu2020triple} identify easy examples based on confidence of the highest class prediction. 
% So easier examples have more confident predictions.
We note that, these methods focus on early exiting \textbf{only} during inference.
In contrast, to the best of our knowledge \method{} is the first work that employs early exiting during \textbf{training} to \textit{learn} the notion of input-hardness. 
Our extensive experiments suggest this new training approach significantly boosts classification accuracy in the long-tailed setting. 
\section{ELF: Learning Input-Hardness For Long Tailed Classification}
In Section~\ref{subsec:intuition} we begin by providing some intuition behind \method{}; then in Section~\ref{subsec:exit_training} and Section~\ref{subsec:exit_inference} we present the 
technical details of \method{} during training and inference, respectively.

\subsection{Input-Hardness Intuition}\label{subsec:intuition}

\begin{wrapfigure}{r}{.35\textwidth}
\vspace{-5mm}
\centering
\includegraphics[width=0.35\textwidth]{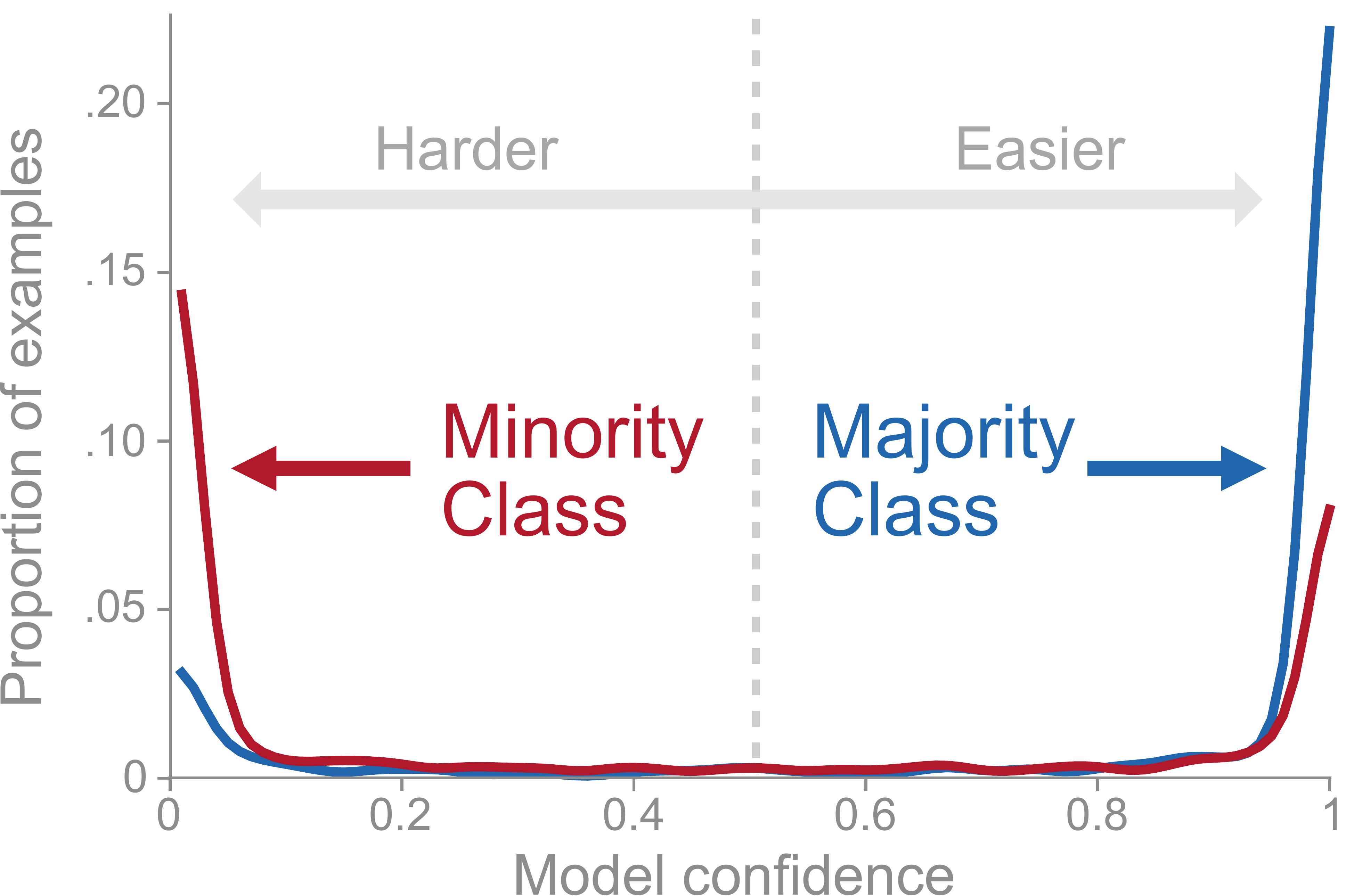}
\vspace{-5mm}
\caption{Proportion of examples in class vs. prediction confidence.}
\label{fig:majority_vs_minority}
\vspace{-5mm}
\end{wrapfigure}

We hypothesize that within both the majority and minority classes some examples are easier than others. 
Consequently, not every example in the minority class needs to be equally upweighted; and not every example in the majority class needs to be equally downweighted.
In order to verify our intuition, we determine what proportion of the rare classes get a high confidence prediction ($\geq0.9$) and what proportion of the majority classes get a low confidence prediction ($\leq$0.1).
Figure~\ref{fig:majority_vs_minority} plots the prediction confidence versus the proportion of examples (in the class) obtaining that confidence from the CIFAR-10 LT dataset.

As expected, many examples from the minority class are classified with low confidence while many examples from the majority class are predicted with near certainty. 
However, confirming our hypothesis, a considerable proportion of the minority class examples obtain a high confidence prediction and vice versa. 
It is precisely this subset of examples---\textit{low confidence majority} and \textit{high confidence minority}---that \method{} impacts the most.
In particular, \method{} increases the loss contribution of low confidence \textit{majority} while retaining the original loss contribution for high confidence \textit{minority}.
This enables a fine-grained, hardness aware approach to loss reweighitng.

\textbf{Notation.}\label{subsec:notation}
We denote an input example as $\pmb{X}_i$ with corresponding label $y_i$ that come from a dataset $\mathcal{D} = \{(\pmb{X}_1,y_1),...,(\pmb{X}_n,y_n)\}$.
The number of training examples in class $j\in \altmathcal{C}$ are denoted by $n_j$, and the total number of training examples is $n = \sum_{j=1}^c n_j$.
Without loss of generality, we sort the classes in descending order of cardinality so that $n_1 \geq...\geq n_{c}$, where $n_1 \gg n_c$ since we operate in the long-tailed setting.  
Our goal is to learn a neural network
$f: \pmb{X} \rightarrow \pmb{z}$ that maps an input space $\pmb{X}$ to a vector of class prediction scores $\pmb{z}=[z_1,...,z_c]^\top$, where $z_i \in [0, 1] \text{ }$.
The neural network is parametrized by weights $\pmb{\theta}^k$ up to the $k^{th}$ exit.
Thus the prediction for input $\pmb{X}_i$ at the $k^{th}$ exit is $\pmb{z}_i^{(k)} = f(\pmb{X}_i; \pmb{\theta}^k)$ where $\pmb{z}_i^{(k)}$ is the prediction confidence over $c$ classes. The confidence for the $j^{th}$ class is obtained through indexing as $\pmb{z}_i^{(k)}[j]$.
Throughout the paper, we use capital bold letters for matrices (e.g., $\pmb{A}$) 
and lower-case bold letters for vectors (e.g., $\pmb{a}$).

\subsection{Early-Exiting During Training}
\label{subsec:exit_training}
We present the idea of training time early exiting that lies at the core of \method.
As shown in Figure~\ref{fig:crown}, \method{} augments a backbone neural network with $K$ auxiliary classifier branches\footnote{we use the terms auxiliary branches and exits interchangeably}. 
During training, each input $(\pmb{X}_i, y_i)\in \altmathcal{D}$ propagates through all auxiliary exits sequentially until it satisfies the following exit criterion---an example exits only when it is predicted \textit{correctly} and with a \textit{high confidence}. 
More formally, the exit criterion $g^{(k)}_i$ for input $\pmb{X}_i$  at exit $k$ is

\begin{equation}
g^{(k)}_i=\left\{
  \begin{array}{@{}ll@{}}
    1, & \text{if}\  \operatorname*{argmax}(\pmb{z}_i^{(k)}) = y_i \text{ and } \pmb{z}_i^{(k)}[y_i] > t^{(k)}\\
    0, & \text{otherwise}
  \end{array}\right.
  \label{eq:early_exit_criterion_training}
\end{equation} 

Here $t^{(k)}$ is the training time threshold at exit $k$. For simplicity, we chose the same value 
of $t^{(k)}$ for all exits.
By construction, our exit-criterion filters out easy examples early on thereby
freeing model capacity for harder examples. 
Conversely, the harder examples do not satisfy the exit criterion and accumulate additional loss at each exit. 
Thus, the overall loss contributed by input $\pmb{X}_i$ is 

\begin{equation}
\label{eq:early_exiting_general}
    \mathcal{L}_{\method}(\pmb{X}_i, y_i) = \sum_{k\in [1, ..., k^{(e)}_{i}]}\mathcal{L}^{(k)}\left(\pmb{z}^{(k)}_i, y_i\right) \text{ , where }k^{(e)}_{i} = \operatorname*{argmin}_{j \in \{1,2,\ldots,K\}} \left(g_i^{(j)}>0\right)
\end{equation}

Here $K$ is the total number of exits, and $k^{(e)}_{i}$  denotes the first exit where example $\pmb{X}_i$ exits by satisfying the exit criterion in Equation~(\ref{eq:early_exit_criterion_training}).
In other words, the \method{} framework aggregates loss from each auxiliary branch until the example exits. 
In contrast, prior work~\cite{teerapittayanon2016branchynet, Huang2018MultiScaleDN} do not perform train time early exiting and aggregates loss from \textbf{every} exit.
We believe that by allowing easy examples to exit during training, we can shift the model's attention to harder examples.
by increasing their loss contribution.
We believe that this difference---\textbf{training time} early-exiting---is essential for increasing the loss contribution of hard examples.

It is important to note that the \method{} loss in Equation~(\ref{eq:early_exiting_general}) is agnostic to the exact instantiation of $\mathcal{L}^{(k)}$ at exit $k$. 
In paricular, $\mathcal{L}^{(k)}$ can be replaced by any loss function useful for class imbalance, including: weighted cross-entropy \cite{cui2019class}, Focal \cite{lin2017focal}, LDAM \cite{cao2019learning} or any combination thereof. 
In practice, we observe consistent improvements with both weighted cross entropy and the recently proposed LDAM loss. 
When using class weighted cross-entropy at each exit, \method{} loss is described as follows

\begin{equation}
    \mathcal{L}_{\method}^{\texttt{CE}}(\pmb{X}_i, y_i) = \sum_{k\in [1, ..., k_i^{(e)}]} \pmb{w}_{y_i} log\left(\frac{exp(\pmb{z}_i^{(k)}[y_i])}{\sum_{j=1}^{C} exp(\pmb{z}_i^{(k)}[j])}\right)
\end{equation}

where $\pmb{w}_{y_i}$ refers to the class specific weight. 
Prior work has proposed various strategies to set class weight $\pmb{w}_{y_i}$ based on inverse class frequency~\cite{huang2016learning, wang2017learning}, inverse square root frequency~\cite{mikolov2013distributed, mahajan2018exploring} and effective weighting~\cite{cui2019class}. 
We leverage the weighting strategy from \cite{cui2019class} that sets $\pmb{w}_c = \frac{1-\beta}{1-\beta^{n_c}}$, where $n_c$ is the number of samples in class $c$ and $\beta$ is a hyperparameter with typical values between $\{0.999, 0.9999\}$. 
% \scott{should say what this does}
When using LDAM~\cite{cao2019learning} at each exit, the \method{} loss can be described as follows

\begin{equation}
\begin{split}
    \mathcal{L}_{\method}^{\texttt{LDAM}}(\pmb{X}_i, y_i) & = \sum_{k\in [1, ..., k_i^{(e)}]} \pmb{w}_{y_i} log\left(\frac{exp(\pmb{z}_i^{(k)}[y_i] - \bigtriangleup_{y_i})}{ exp(\pmb{z}_i^{(k)}[y_i] - \bigtriangleup_{y_i}) + \sum_{j \neq y_i} exp(\pmb{z}_i^{(k)}[j])}\right) \\ \\
    & \text{and } \bigtriangleup_{y_i} = \frac{C}{n_{y_i}}
\end{split}
\end{equation}

where $\bigtriangleup_{y_i}$ is the per class margin that ensures rare classes get a larger margin.
It is determined through a hyperparameter $C$ and the number of examples $n_{y_i}$ in class $y_i$. 
We select $C$ such that the largest margin for any class is capped at 0.5 \cite{cao2019learning}.

A desirable outcome of training time early-exiting is that harder examples contribute a higher average loss than easier examples.  
Formally, we define this through the increasing loss property:

\begin{property} (Increasing Loss Property) If $D_{k}$ denotes the set of examples exiting at exit $k$ then
$\mathop{\mathbb{E}}_{(\pmb{X}_i, y_i)\in D_{1}}[\mathcal{L}_{\method}(\pmb{X}_i, y_i)] < \mathop{\mathbb{E}}_{(\pmb{X}_i, y_i)\in D_{2}}[\mathcal{L}_{\method}(\pmb{X}_i, y_i)] < ... < \mathop{\mathbb{E}}_{(\pmb{X}_i, y_i)\in D_{k}}[\mathcal{L}_{\method}(\pmb{X}_i, y_i)]$. 
\label{prop:increasing_loss_property}
\end{property}

\begin{wrapfigure}{r}{.3\textwidth}
\centering
\vspace{-4mm}
    \includegraphics[width=0.3\textwidth]{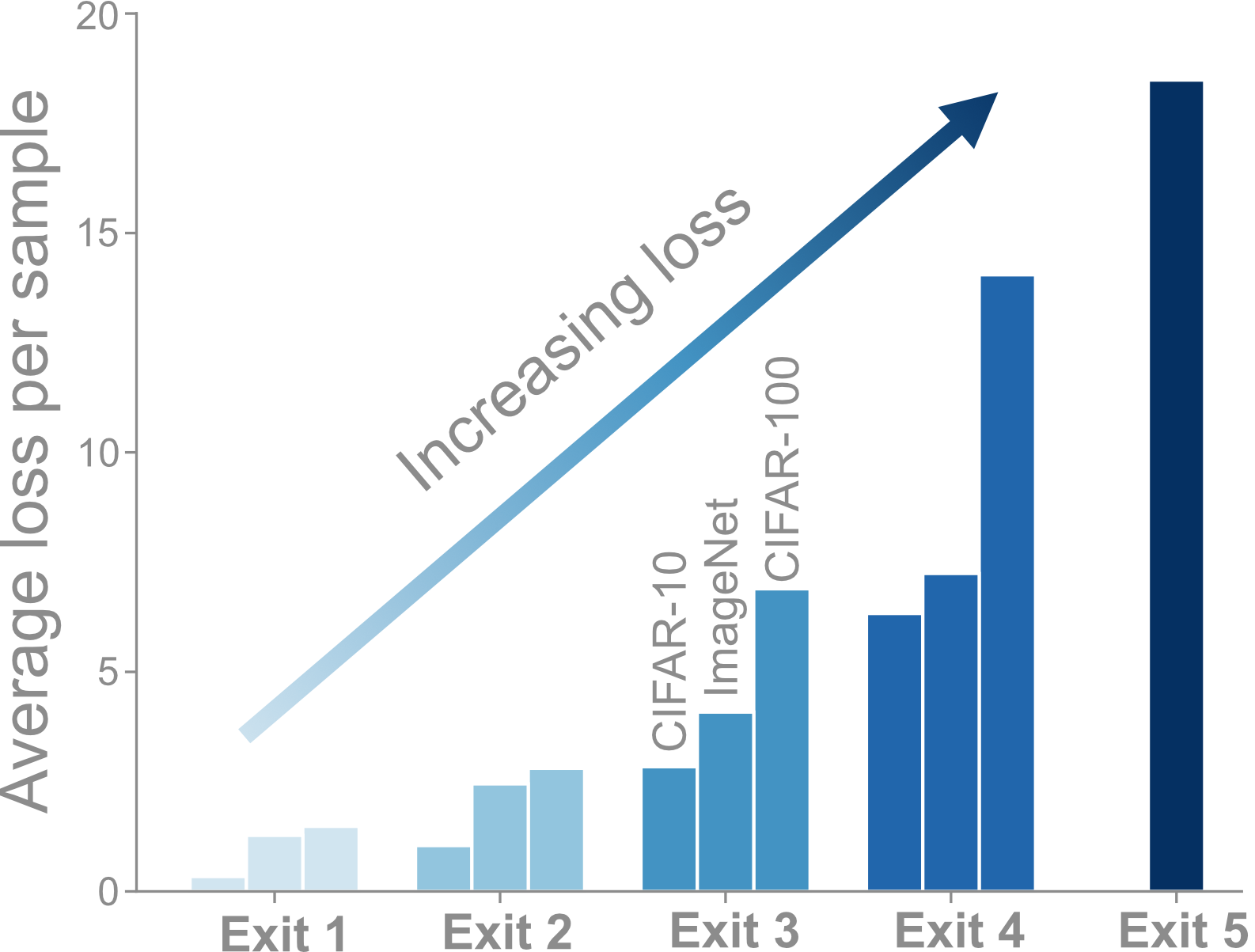}
    \caption{Average per sample loss for images exiting at different exits on three datasets. The increasing trend validates Property~\ref{prop:increasing_loss_property}.
    }
    \label{figure:loss}
\vspace{-7mm}
\end{wrapfigure}

This property enables the neural network to focus on harder examples which contribute a higher expected loss.
In Figure~\ref{figure:loss}, we plot the average training loss contributed by examples exiting at different auxillary branches. 
The increasing trend of average loss across exits validates the increasing loss property.

\subsection{Early-Exiting During Inference}
\label{subsec:exit_inference}
Training with \method{} loss enables a neural network to learn a notion of input-hardness. 
During inference, this can be leveraged to early-exit examples from both the minority and majority classes based on hardness.  
Formally, the inference time early-exit criterion $h^{(k)}_i$ (for an input $\pmb{X}_i$) at the $k^{th}$ exit is a relaxed version of Equation~(\ref{eq:early_exit_criterion_training}) and defined below:

\begin{equation}
    h^{(k)}_i=\left\{
  \begin{array}{@{}ll@{}}
    1, & \text{if}\  \operatorname*{argmax}(\pmb{z}_i^{(k)}) > s^{(k)}\\
    0, & \text{otherwise}
  \end{array}\right.
    \label{eq:early_exit_criterion_inference}
\end{equation}

Here $\pmb{z}_i^{(k)}$ is the prediction vector for the input $\pmb{X}_i$ obtained at exit $k$, and $s^{(k)}$ is the inference time threshold, which for simplicity we set to be the same across all exits.
Our inference time exit criterion in Equation~(\ref{eq:early_exit_criterion_inference}) exits examples based on prediction confidence which is in line with prior work \cite{huang2017multi, Phuong_2019_ICCV}. 
Using this criterion, we obtain the prediction vector $\pmb{z}_i$ for input $\pmb{X}_i$  as

\begin{equation}
    \pmb{z}_i = \pmb{z}_i^{(k^{(e)}_{i})} \text{ , where }k^{(e)}_{i} = \operatorname*{argmin}_{j \in \{1,...,K\}} \left(h_i^{(j)}>0\right)
\end{equation}
Here $k^{(e)}_{i}$ denotes the first exit where the inference time exit-criterion of Equation~(\ref{eq:early_exit_criterion_inference}) holds. 
We note that the prediction vector $\pmb{z}_i$ is determined by both the input $\pmb{X}_i$ and the inference exit threshold $s^{(k)}$. Furthermore, reducing $s^{(k)}$ causes more examples to early exit, which in turn leads to a reduction in FLOPS. 
Thus, given a \textit{single} model trained using \method{} (Equation~(\ref{eq:early_exiting_general})), varying  $s^{(k)}$ offers a way to dynamically generate a family of models with different compute budgets.

\section{Experiments}
In Section~\ref{subsec:experimental_setup}, we begin by discussing the experimental setup including: (i) evaluated datasets, (ii) model and training configuations, and (iii) loss function setup.
Next, in Section~\ref{subsec:classification_performance} we extensively analyze \method{}'s long-tailed performance and disect its ability to improve upon the state-of-the-art.

\subsection{Experimental Setup}\label{subsec:experimental_setup}

\begin{figure*}[t!]
    \includegraphics[width=\textwidth]{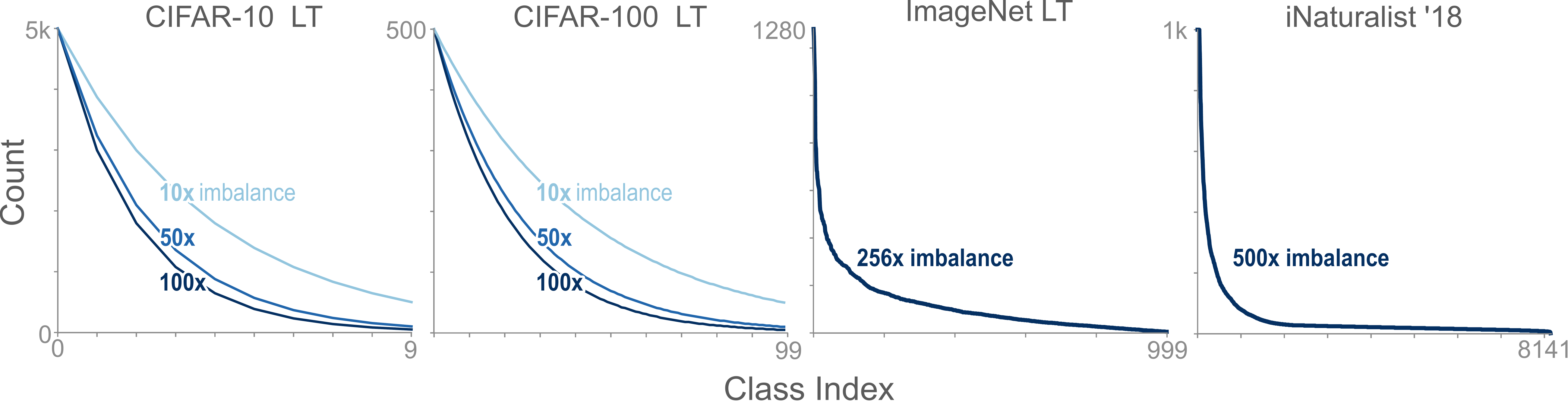}
    \caption{
    Training distribution for each dataset. 
    On CIFAR-10 and CIFAR-100 we evaluate 3 levels of data imbalance: $10$x, $50$x and $100$x. 
    ImageNet LT and iNaturalist'18 have imbalance ratios of $256\times$ and $500\times$, respectively. For long-tailed datasets, a majority of classes only have a few examples.  
    }
    \label{fig:imbalance}
    % \vspace{-1cm}
\end{figure*}

\textbf{Datasets.}
We conduct our evaluation on four long-tailed datasets: CIFAR-10 LT, CIFAR-100 LT ~\cite{cui2019class}, ImageNet LT~\cite{liu2019large} and iNaturalist'18~\cite{van2018inaturalist}. 
For the first three datasets, the training split is obtained by subsampling from their balanced versions: CIFAR-10~\cite{krizhevsky2009learning}, CIFAR-100~\cite{krizhevsky2009learning} and Imagenet 2012~\cite{russakovsky2015imagenet}. 
In case of the CIFAR LT datasets, we consider three levels of imbalance, $10\times$, $50\times$ and $100\times$, which is defined as the ratio between the number of samples in the largest and the smallest classes. 
The ImageNet LT training split consists of 115.8k images from 1,000 classes with largest and smallest classes containing 1,280 and 5 images respectively. 
The iNaturalist'18 dataset is a naturally imbalanced dataset containing 437,513 training images from 8,142 categories with a test set of 24,426 images. 
Figure~\ref{fig:imbalance} highlights the training data distribution for all four datasets. 
We note that for each dataset, the validation and test sets are balanced across classes and thus top-1 accuracy serves as a common metric of comparison. 
See the Appendix for additional details on dataset construction.

\textbf{Backbone models \& training configurations.}\label{subsec:architecture}
We evaluate several models from the ResNet and DenseNet families. To obtain the \method{} models, we attach auxiliary classifier branches before each residual/dense block (see Appendix for details). 
On CIFAR datasets, we train all ResNet-32 models for 200 epochs using SGD with an initial learning rate of $0.1$ decreased by $0.01$ at epochs $160$ and $180$~\cite{cao2019learning,cui2019class}. 
The weight decay is $2\times 10^{-4}$. 
On ImageNet LT and iNaturalist'18 we train all ResNet-50 and DenseNet-169 models for 100 epochs using SGD with an initial learning rate of 0.1 decreased by 0.1 at epochs 60 and 80. 
The weight decay is $2\times 10^{-4}$. 
All models use a linear warmup schedule for the first 5 epochs to avoid initial overfitting in the imbalanced setting \cite{cui2019class, cao2019learning}.

\begin{figure}[b]
    \centering
    \includegraphics[width=0.65\linewidth]{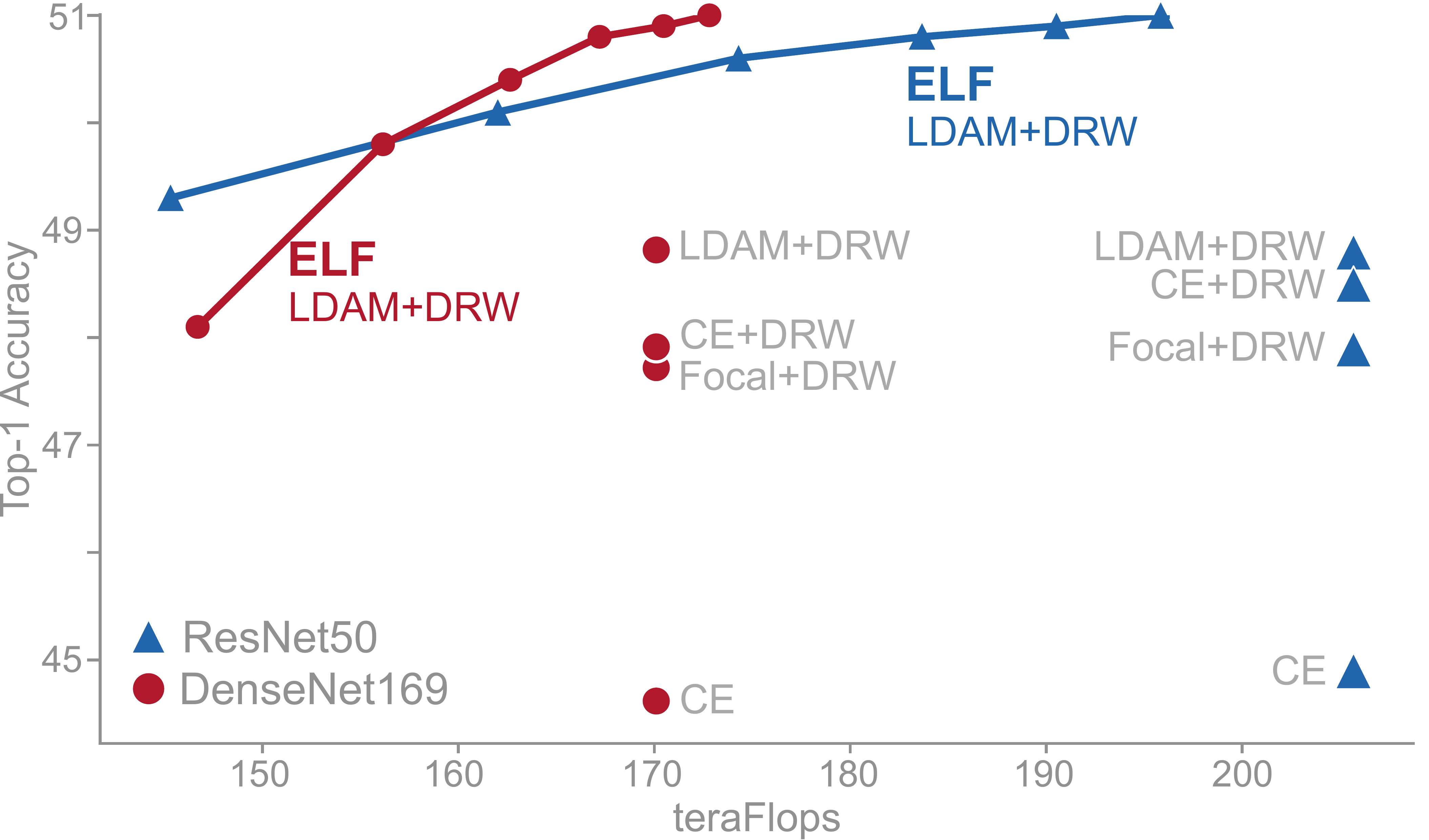}
    \caption{By varying the inference threshold $s^{(k)}$, \method{} enables on-the-fly model selection based on the available compute budget (red, blue curves). 
    Each point depicts a ResNet-50 or DenseNet-169 model trained on ImageNet LT for 100 epochs. Observe that the models generated through \method{} achieve more than $3\%$ accuracy gains over any other method while using similar or fewer FLOPS. 
    }
    \label{fig:inference_ablation}
\end{figure}

\textbf{Implementation.} 
\method{} is constructed from three \textit{key} components---(1) a class reweighting strategy, (2) an exit loss function and (3) a reweighting schedule.
For class reweighting, we weight each example of class $c$ according to it's effective number $\frac{1-\beta}{1-\beta^{n_c}}$, where $n_c$ is the number of images in class $c$ \cite{cui2019class}.
For the exit loss, we consider two variations with \method{}---at each exit we use either cross entropy  loss (CE) or label distribution aware margin loss (LDAM)~\cite{cao2019learning}. 
These are referred to as \method{}{\scriptsize (CE)} and \method{}{\scriptsize (LDAM)}.
Finally, for the reweighting schedule we use the per-dataset delayed reweighting (DRW) scheme introduced in \cite{cao2019learning}.
All experiments are conducted in PyTorch 1.0 using an Nvidia DGX-1 containing eight V100 GPUs and 512GB of RAM.

For \method{}, we set the training and exit thresholds $t^{(k)}, s^{(k)}$ based on the exit loss type.
\method{}{\scriptsize (CE)} uniformly sets $t^{(k)}$ = $0.9$ for all exits, while \method{}{\scriptsize (LDAM)} sets $t^{(k)}$ = $2/|c|$, where $|c|$ is the number of classes in the dataset. 
The inference exit threshold $s^{(k)}$ is identified through a line search.
On large datasets, we observe that \method{\scriptsize (LDAM)} takes more epochs to converge, therefore we provide results for both 100 and 200 epochs.
To measure the performance of \method{}, we compare against three strong baselines: CE \cite{cui2019class}, Focal \cite{lin2017focal} and LDAM \cite{cao2019learning}, each reusing the same class reweighting and delayed reweighting schedule discussed above. 
For Focal loss we set $\gamma$ = $0.5$ \cite{lin2017focal}, and for LDAM we set $C$ such that the maximum margin is 0.5 \cite{cao2019learning}.
See Appendix for additional implementation details.

\subsection{Evalution on Long-Tailed Classification.}\label{subsec:classification_performance}
We begin by 
highlighting \method{}'s ability to \textit{train once and generate a family of models}.
Next, we discuss and analyze the performance of \method{} on the task of long-tailed classification.

\textbf{Generating a family of models along the Accuracy-FLOP curve.} In Figure~\ref{fig:inference_ablation} we observe the Accuracy-FLOP trade-off for \method{} models trained on ImageNet LT. 
For \method{}, the models lie along a curve that offers a favorable accuracy-efficiency tradeoff. 
The models on this curve are obtained by training \textbf{once} using a fixed training threshold $t^{(k)}$ while linearly varying the inference threshold $s^{(k)}$. 
This leads to a family of \method{} models that enables on-the-fly model selection based on a compute budget.
We note that models trained without \method{} stack vertically since they consume the same number of inference FLOPS.

\textbf{Evaluating classification accuracy.}
In Table~\ref{tab:large_scale_results} \textit{All} column, we compare the top-1 accuracy of \method{} to each baseline configuration on ImageNet LT and iNaturalist'18 datasets.
Table~\ref{tab:cifar_experiments} presents similar analysis on CIFAR-10/100 LT. 
In both tables we observe that \method{} consistently improves the state-of-the-art accuracy. 
Moreover the margin of improvement is higher for increasing levels of imbalance (see Table~\ref{tab:cifar_experiments}). 
One interesting insight is that \method{} improves performance independant of loss type at each exit.
Specifically, we see consistent improvements while going from CE~\cite{cui2019class} to \method{}{\scriptsize (CE)} and from LDAM~\cite{cao2019learning} to \method{}{\scriptsize (LDAM)}.

% \begin{table}[]
\begin{table*}[b!]
\centering
\setlength{\tabcolsep}{4pt}
\begin{tabular}{lrrrarrra}

\toprule
& \multicolumn{4}{c}{Imagenet-LT} & \multicolumn{4}{c}{iNaturalist '18} \\ 
\cmidrule(l){2-5}\cmidrule(l){6-9}
\textbf{}                        & Many & Med & Few & All & Many & Med & Few & All \\
\hline
CE                     &       63.8      &        38.5         &      13.6        &        44.6 \scriptsize{(\textcolor{black}{0\%})}   & \textbf{72.7} & 63.8 & 58.7 & 62.7 \scriptsize{(\textcolor{black}{0\%})} \\
BBN$^{\dagger}$\scriptsize{}~\cite{zhou2019bbn}                     &      -        &       -         &      -       &       - & - & - & - & 69.6 \scriptsize{(\textcolor{red}{+100\%})} \\
% % \midrule
CRT$^{\dagger}$\scriptsize{}~\cite{kang2019decoupling}                     &       58.8        &        44.0         &      26.1        &        47.3  \scriptsize{(\textcolor{black}{0\%})}& 69.0 & 66.0 & 63.2 & 65.2 \scriptsize{(\textcolor{black}{0\%})}\\

LWS$^{\dagger}$\scriptsize{}~\cite{kang2019decoupling}                     &       57.1        &        45.2         &      29.3        &        47.7 \scriptsize{(\textcolor{black}{0\%})}& 65.0 & 66.3 & 65.5 & 65.9 \scriptsize{(\textcolor{black}{0\%})}\\
$\tau$-norm$^{\dagger}$\scriptsize{}~\cite{kang2019decoupling}          &       56.6        &        44.2         &      27.4        &        46.7 \scriptsize{(\textcolor{black}{0\%})}& 65.6 & 65.3 & 65.9 & 65.6 \scriptsize{(\textcolor{black}{0\%})} \\%\vspace{2mm}\\
\hline

CE+\scriptsize{DRW}~\cite{cui2019class}                  &       60.3        &        45.2         &       27.0         &       48.5 \scriptsize{(\textcolor{black}{0\%})} & 67.1 & 66.4 & 65.6 & 66.1  \scriptsize{(\textcolor{black}{0\%})}\\
Focal+\scriptsize{DRW}~\cite{lin2017focal}               &       59.5        &        44.6         &       27.0       &       47.9   \scriptsize{(\textcolor{black}{0\%})}& 66.1 & 66.0 & 64.3 & 65.4  \scriptsize{(\textcolor{black}{0\%})} \\

LDAM+\scriptsize{DRW}~\cite{cao2019learning}                &       61.1       &        44.7         &      28.0        &        48.8   \scriptsize{(\textcolor{black}{0\%})}& 70.0 & 67.4 & 66.1 & 67.1\ \scriptsize{(\textcolor{black}{0\%})}  \\%\vspace{2mm}\\
\hline
\method{}{\scriptsize (CE) + DRW} \tiny{(100 epochs)}  & 60.7 & 45.5  & 27.7 &  48.9 \scriptsize{(\textcolor{blue}{\textbf{-20.7}\%})}& 67.4 & 66.3 & 65.1 & 66.0 \scriptsize{(\textcolor{blue}{\textbf{-13.5}\%})} \\
\method{}{\scriptsize (LDAM) + DRW} \tiny{(100 epochs)}    &  64.0      &      46.8        &        27.7         &       50.8 \scriptsize{(\textcolor{blue}{\textbf{-7.4}\%})}   & 72.3 & 68.8 & 65.6 & 67.9  \scriptsize{(\textcolor{blue}{\textbf{-9.0}\%})}  \\
\method{}{\scriptsize (LDAM)+ DRW} \tiny{(200 epochs)}  &  \textbf{64.3} &  \textbf{47.9} & \textbf{31.4} & \textbf{52.0} \scriptsize{(\textcolor{blue}{\textbf{-10.0}\%})} & \textbf{72.7}  & \textbf{70.4} & \textbf{68.3} & \textbf{69.8} \scriptsize{(\textcolor{blue}{\textbf{-12.6}\%})}  \\
\bottomrule
% EEL-\scriptsize{}LDAM+DRW   &   \textbf{64.3}      &      \textbf{47.9}         &     \textbf{31.4}           &    \textbf{52.0}  \\ \bottomrule       
\end{tabular}
\caption{Top-1 accuracy for ResNet-50 trained on Imagenet LT and iNaturalist’18 datasets. The overall accuracy (All column) is decomposed into three splits corresponding to \textit{many}, \textit{medium} and \textit{few} shot settings. Numbers in parenthesis indicate the FLOPS expended by each method relative to the baseline model CE (i.e., more \textcolor{blue}{\textit{negative}} means more savings, thus better). \method{} consistently improves accuracy while expending fewer FLOPS. $^\dagger$Original results from the referenced paper. 
% \rd{INAT : Many=843, Med=4076, Few=3224] }
}
\label{tab:large_scale_results}
\end{table*}
\begin{table*}[t]
\setlength{\tabcolsep}{2pt}
\centering
\begin{tabular}{lllllll}
\toprule

& \multicolumn{3}{c}{CIFAR-10 Long Tailed} & \multicolumn{3}{c}{CIFAR-100 Long Tailed} \\ 

\cmidrule(l){2-4}\cmidrule(l){5-7}

\textbf{Method} & 100 & 50 &  10 & 100 & 50 & 10 \\ 

\midrule

% CE~\cite{he2016deep}  & 29.6 & 25.2 &  13.6 & 61.7 & 56.1 &  44.3 \\%[0.5ex]
CE  & 70.4 \scriptsize{(\textcolor{black}{0\%})} & 74.8 \scriptsize{(\textcolor{black}{0\%})} &  16.4 \scriptsize{(\textcolor{black}{0\%})} & 28.3 \scriptsize{(\textcolor{black}{0\%})} & 43.9 \scriptsize{(\textcolor{black}{0\%})} &  55.7 \scriptsize{(\textcolor{black}{0\%})} \\%[0.5ex]
BBN$^{\dagger}$\scriptsize{}~\cite{zhou2019bbn}  & \textbf{79.8} \scriptsize{(\textcolor{red}{+100\%})} & 82.2 \scriptsize{(\textcolor{red}{+100\%})} &  \textbf{88.3} \scriptsize{(\textcolor{red}{+100\%})} & 42.5 \scriptsize{(\textcolor{red}{+100\%})} & 47.0 \scriptsize{(\textcolor{red}{+100\%})} &  \textbf{59.1}  \scriptsize{(\textcolor{red}{+100\%})} \\%[0.5ex]
% Focal~\cite{lin2017focal}  & 29.6 & 23.3 &  13.3 & 61.6 & 55.7 &  44.2 \\%[0.5ex]
Focal$^{\dagger}$\scriptsize{}~\cite{lin2017focal}  & 70.4 \scriptsize{(\textcolor{black}{0\%})} & 76.7 \scriptsize{(\textcolor{black}{0\%})} &  86.7 \scriptsize{(\textcolor{black}{0\%})} & 28.4 \scriptsize{(\textcolor{black}{0\%})} & 44.3 \scriptsize{(\textcolor{black}{0\%})} &  55.8 \scriptsize{(\textcolor{black}{0\%})}\\%[0.5ex]
% Mixup~\cite{zhang2017mixup}  & 26.9 & 22.2 &  12.9 & 60.5 & 55.0 &  42.0 \\%[0.5ex]
Mixup$^{\dagger}$\scriptsize{}~\cite{zhang2017mixup}  & 73.1 \scriptsize{(\textcolor{black}{0\%})} & 77.8 \scriptsize{(\textcolor{black}{0\%})} &  87.1 \scriptsize{(\textcolor{black}{0\%})} & 39.5 \scriptsize{(\textcolor{black}{0\%})} & 45.0 \scriptsize{(\textcolor{black}{0\%})} &  58.0 \scriptsize{(\textcolor{black}{0\%})}\\%[0.5ex]
% Manifold Mixup~\cite{verma2018manifold}  & 27.0 & 22.0 &  13.0 & 61.7 & 56.9 &  43.5 \\%[0.5ex]
Manifold Mixup$^{\dagger}$\scriptsize{}~\cite{verma2018manifold}  & 73.0 \scriptsize{(\textcolor{black}{0\%})}& 78.0 \scriptsize{(\textcolor{black}{0\%})}&  87.0 \scriptsize{(\textcolor{black}{0\%})}& 38.3 \scriptsize{(\textcolor{black}{0\%})}& 43.1 \scriptsize{(\textcolor{black}{0\%})}&  56.5\scriptsize{(\textcolor{black}{0\%})}\\%[0.5ex]
% BBN ($2\times$)  & \textbf{20.2} (\textcolor{red}{+100}) & 17.8 (\textcolor{red}{+100}) &  \textbf{11.7} (\textcolor{red}{+100}) & 57.5 (\textcolor{red}{+100}) & 53.0 (\textcolor{red}{+100}) &  40.9  (\textcolor{red}{+100}) \\%[0.5ex]

\midrule
% CE-DRW~\cite{cui2019class} & 23.7 & 20.0  & 12.4 & 58.6 & 54.0 &  41.7 \\%[0.5ex]
CE+\scriptsize{DRW}~\cite{cui2019class} & 76.3 \scriptsize{(\textcolor{black}{0\%})} & 80.0 \scriptsize{(\textcolor{black}{0\%})} & 87.6 \scriptsize{(\textcolor{black}{0\%})}& 41.4 \scriptsize{(\textcolor{black}{0\%})}& 46.0 \scriptsize{(\textcolor{black}{0\%})}&  58.3 \scriptsize{(\textcolor{black}{0\%})}\\%[0.5ex]
% Focal-DRW~\cite{lin2017focal} & 25.4 & 20.6 &  12.7 & 60.6 & 54.7 &  42.5 \\%[0.5ex]
Focal+\scriptsize{DRW}~\cite{lin2017focal} & 74.6 \scriptsize{(\textcolor{black}{0\%})}& 79.4 \scriptsize{(\textcolor{black}{0\%})}&  87.3 \scriptsize{(\textcolor{black}{0\%})}& 39.4 \scriptsize{(\textcolor{black}{0\%})}& 45.3 \scriptsize{(\textcolor{black}{0\%})}&  57.5 \scriptsize{(\textcolor{black}{0\%})}\\%[0.5ex]
% LDAM-DRW~\cite{cao2019learning} & 23.0 & 18.6 &  12.4 & 58.0 & 53.4 &  41.3 \\%[0.5ex]
LDAM+\scriptsize{DRW}~\cite{cao2019learning} & 77.0 \scriptsize{(\textcolor{black}{0\%})}& 81.4 \scriptsize{(\textcolor{black}{0\%})}&  87.6 \scriptsize{(\textcolor{black}{0\%})}& 42.0 \scriptsize{(\textcolor{black}{0\%})}& 46.6 \scriptsize{(\textcolor{black}{0\%})}&  58.7 \scriptsize{(\textcolor{black}{0\%})}\\%[0.5ex]
\midrule
% EEL-DRW  & 21.9 (\textcolor{green}{-15.1}) & \textbf{17.6} (\textcolor{green}{-21.4}) &  12.0 (\textcolor{green}{-19.6}) & \textbf{56.9} (\textcolor{green}{-0.01}) & \textbf{52.5}  (\textcolor{green}{-2.39}) & \textbf{41.3} (\textcolor{green}{-9.8}) \\%[0.5ex]
\method{}{\scriptsize (CE)}+\scriptsize{DRW}  & 76.8 \scriptsize{(\textcolor{blue}{\textbf{-31.9}\%})} & 80.8 \scriptsize{(\textcolor{blue}{\textbf{-28.8}\%})} &  87.6 \scriptsize{(\textcolor{blue}{\textbf{-26.4}\%})} & 42.5 \scriptsize{(\textcolor{blue}{\textbf{-11.6}\%})} & 47.1 \scriptsize{(\textcolor{blue}{\textbf{-11.5}\%})} & 58.7 \scriptsize{(\textcolor{blue}{\textbf{-9.8}\%})} \\

\method{}{\scriptsize (LDAM)}+\scriptsize{DRW}  & 78.1 \scriptsize{(\textcolor{blue}{\textbf{-15.1}\%})} & \textbf{82.4} \scriptsize{(\textcolor{blue}{\textbf{-21.4}\%})} &  88.0 \scriptsize{(\textcolor{blue}{\textbf{-19.6}\%})} & \textbf{43.1} \scriptsize{(\textcolor{blue}{\textbf{-0.01}\%})} & \textbf{47.5}  \scriptsize{(\textcolor{blue}{\textbf{-2.39}\%})} & 58.9 \scriptsize{(\textcolor{blue}{\textbf{-1.9}\%})} \\%[0.5ex]
% Flops Saving (in \%) & 15.1 &  21.4 &  19.6 & 0.01 & 2.39 &  9.8  \\ %[0.5ex]

% Exit-Loss & \scriptsize{LDAM} &  \scriptsize{LDAM} &  \scriptsize{LDAM} & \scriptsize{LDAM} & \scriptsize{LDAM} &  \scriptsize{CE}  \\ 

% Flops Saving (in \%) & 31.9 &  28.8 &  26.4 & 11.6 & 11.5 &  9.8  \\ 
% \method-D (LDAM)  & \textbf{21.9} & \textbf{17.6} &  \textbf{12.0} & \textbf{56.9} & \textbf{52.5} & 43.0 \\[0.5ex]
% Flops Saving (in \%) & 15.1 &  21.4 &  19.6 & 0.00 & 2.39 &  1.50  \\ 
% \method-D  & \textbf{22.9} & \textbf{18.5} & \textbf{13.8} & \textbf{11.6} & \textbf{57.8} & \textbf{53.0} & \textbf{45.9} & \textbf{40.8} \\[0.5ex]
% Flops Saving (in \%) & 23.0 &  21.5 & 17.2 & 21.4 & 3.4 & 15.4 & 3.6 & 9.9  \\ 
\bottomrule     
\end{tabular}
\caption{Top-1 accuracy for ResNet-32 models trained on long tailed CIFAR-10 and CIFAR-100 datasets. 
Numbers in parentheses indicate the FLOPS expended relative to the baseline model CE (i.e., more \textcolor{blue}{\textit{negative}} means more savings, thus better). Notice that BBN outperforms \method in some scenarios. This is not surprising since it uses double the number of FLOPS compared to all other methods. $^\dagger$These results are referenced from \cite{zhou2019bbn}. 
}.
\label{tab:cifar_experiments}

\end{table*}

\textbf{Dissecting the accuracy improvement.}
To dissect the accuracy of each method, we break down the combined test set into three sets of classes---\textit{Many}, \textit{Medium} and \textit{Few}.
These splits refer to classes containing more than 100 examples as \textit{Many}, classes with 20\textasciitilde{}100 examples as textit{Medium} and classes with less than 20 examples as \textit{Few}~\cite{kang2019decoupling}. 
From Table~\ref{tab:large_scale_results} we observe that \method{\scriptsize (LDAM)} comprehensively outperforms prior methods on all three splits. 
Interestingly, we observe that traditional class imbalance techniques sacrifice accuracy on the majority class in order to improve the medium and few classes.
In contrast, \method{} maintains or improves accuracy across all three class splits.

To ascertain whether the accuracy gains are due to the extra model capacity introduced by exit branches, we calculate the total FLOPS consumed by each model on the test set.  
The relative FLOPS savings with respect to the baseline model (CE) are presented in parenthesis in Tables~\ref{tab:large_scale_results} and \ref{tab:cifar_experiments}. 
\method{} always reduces the FLOPS count, saving up to 20\% FLOPS while improving overall accuracy by over 3\%. 
In contrast, a recent method (BBN \cite{zhou2019bbn}) achieves similar accuracy by using more than double the FLOPS of the baseline CE model.
We ascribe \method's improvement in generalization to the hardness aware learning, which we further discuss below.

\begin{figure}[t!]
    \centering
    \includegraphics[width=0.85\textwidth]{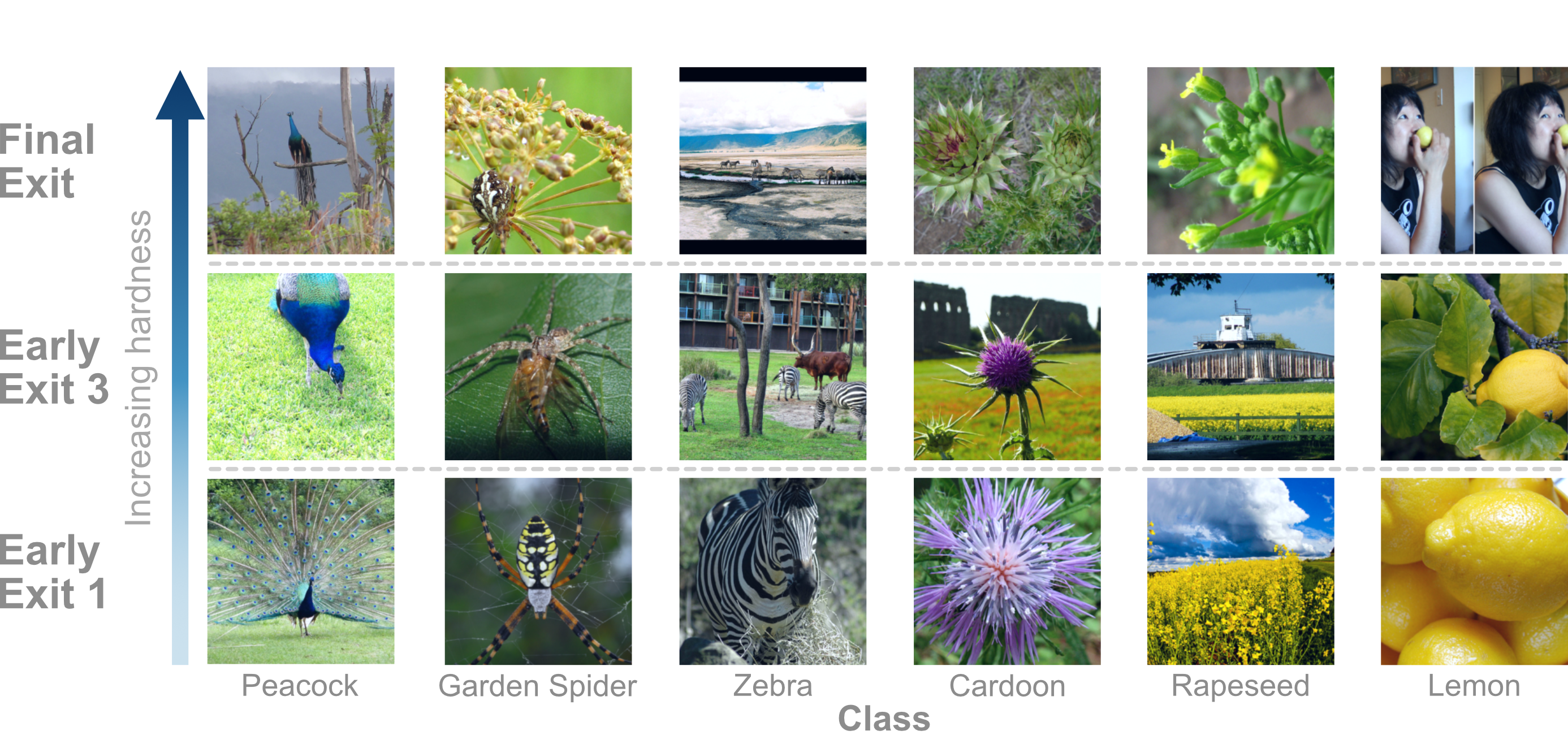}
    \caption{Images exiting from the 1st, 3rd and final exit of the \method{} framework. 
    In the 1st exit we observe easy to classify images. As the exits increase, so does the image hardness.
    }
    \label{fig:image-difficulty}
\end{figure}

\textbf{Visualizing the learned notion of input-hardness.}
In Figure~\ref{fig:image-difficulty} we present a variety of test set images exiting through auxiliary branches in an \method{} ResNet-50 model trained on ImageNet LT. 
For a particular class, we observe that the object of interest is easier to distinguish in images exiting from earlier branches.
% We observe that for a given class, the object of interest in images exiting earlier in the network is easier to distinguish compared to images exiting later.
% We observe that objects of interest in images exiting earlier in the network are easier to distinguish compared to their counterparts that exit from later branches.
% is harder to distinguish and thus require additional cogitive effort to interpret.
For example, consider the lemon class. In the image obtained from exit 1 (column 6, row 3), the object of interest is clearly visible.
% For example, within the lemon class, the object of interest is clearly visible in the image obtained from exit 1 (column 6, row 3).
In comparison, lemon images obtained at exit 3 and (final) exit 5 (column 6, rows 2 and 1), the object of interest is largely obstructed. 
This highlights that \method{} enables a model to learn an intuitive notion of image hardness.

\section{Conclusion}
We identified the notion of \textit{sample-hardness} as a key concept to improve generalization under a long-tailed class distribution. 
To incorporate this notion of hardness in the learning process, we proposed the \method{} framework. 
\method{} is complementary to existing work in long-tailed classification and can readily integrate with existing approaches to improve classification accuracy. 
Extensive evaluations demonstrate that \method{} outperforms existing state-of-the-art techniques while enabling on-the-fly model selection for varying compute budgets.

\section*{Broader Impact}
We describe two scenarios where \method{} can have a high impact.

\textbf{Disease classification.} Long-tailed data distributions arise in many real world use cases.
For example, a typical prediction task on electric health records (EHR) usually involves classifying over 10,000 diseases codes, many of which are rare. 
Obtaining good generalization performance on these rare classes is an extremely challenging problem. 
The largest dataset considered in this paper--iNaturalist'18--- contains 8,142 classes and presents a challenge of comparable complexity. 
On iNaturalist'18, \method{} improves state-of-the-art accuracy by over 2.7\% (compared to LDAM loss~\cite{cao2019learning}) while saving over $12.5\%$ FLOPS during inference. 

\textbf{Edge device deployment.} Another benefit of \method{} is that it enables a model to dynamically vary it's compute footprint during inference.
% This enables any \method{} model to adapt to the available compute budget. 
A real-world use case arises when a model is deployed to an edge device (e.g., smartphone, tablet, embedded system).
As the battery level decreases, an \method{} model can reduce its computational footprint in real time. 

\textbf{Potential weaknesses.} Like many deep learning methods, accurate model performance is dependent on large quantities of labeled data.
Unfortunately, this is not always available in every domain. 
In addition, our proposed method is ethically neutral meaning we need to pay attention to how it is used in practice.

% An AI for healthcare system will need to perform well in all 10,000 classes which can be extremely challenging. 

% \noindent{\bf Potential benefit for fair ML models} The model performance often 
% \noindent {\bf Caveat and potential weakness:}
% require larege training data
\bibliographystyle{unsrt}
\bibliography{main}

\newpage

\section*{Appendix}
\label{sec:appendix}
\appendix

\section{Additional Results on DenseNet-169}
In Table~\ref{tab:densenet_ablation}, we present the top-1 accuracy acheived by a DenseNet-169 model trained on ImageNet LT with various loss functions. 
Similar to our findings with ResNet-50 (see Table~\ref{tab:large_scale_results} in the main paper), we observe that models trained with \method{}{\scriptsize (LDAM)} improve more than $2.5\%$ on top-1 accuracy while consuming fewer FLOPS. Moreover, the accuracy improvement is acheived on the three class splits corresponding to the \textit{Many}, \textit{Med} and \textit{Few} shot settings.

% \begin{table}[]
\begin{table*}[h!]
\centering
\setlength{\tabcolsep}{4pt}
\begin{tabular}{lrrrarrra}

\toprule
& \multicolumn{4}{c}{Imagenet LT} & \multicolumn{4}{c}{iNaturalist '18} \\ 
\cmidrule(l){2-5}\cmidrule(l){6-9}
\textbf{}                        & Many & Med & Few & All & Many & Med & Few & All \\
\hline
CE                     &       63.5      &       38.1      &      14.4      &       44.6 \scriptsize{(\textcolor{black}{0\%})}    & \textbf{73.9} & 64.6 & 58.1 & 63.0 \scriptsize{(\textcolor{black}{0\%})} \\
% BBN$^{\dagger}$\scriptsize{}~\cite{zhou2019bbn}                     &      -        &       -         &      -       &       - & - & - & - & 69.6 \scriptsize{(\textcolor{red}{+100\%})} \\
% % % \midrule
% CRT$^{\dagger}$\scriptsize{}~\cite{kang2019decoupling}                     &       58.8        &        44.0         &      26.1        &        47.3  \scriptsize{(\textcolor{black}{0\%})}& 69.0 & 66.0 & 63.2 & 65.2 \scriptsize{(\textcolor{black}{0\%})}\\

% LWS$^{\dagger}$\scriptsize{}~\cite{kang2019decoupling}                     &       57.1        &        45.2         &      29.3        &        47.7 \scriptsize{(\textcolor{black}{0\%})}& 65.0 & 66.3 & 65.5 & 65.9 \scriptsize{(\textcolor{black}{0\%})}\\
% $\tau$-norm$^{\dagger}$\scriptsize{}~\cite{kang2019decoupling}          &       56.6        &        44.2         &      27.4        &        46.7 \scriptsize{(\textcolor{black}{0\%})}& 65.6 & 65.3 & 65.9 & 65.6 \scriptsize{(\textcolor{black}{0\%})} \\%\vspace{2mm}\\
\hline

CE+\scriptsize{DRW}~\cite{cui2019class}                  &       60.3        &        44.2        &      25.8        &       47.9 \scriptsize{(\textcolor{black}{0\%})} & 68.9 & 67.3 & 65.6 & 66.8  \scriptsize{(\textcolor{black}{0\%})}\\
Focal+\scriptsize{DRW}~\cite{lin2017focal}               &       59.8 & 44.1 & 26.0 & 47.7 \scriptsize{(\textcolor{black}{0\%})}& 68.3&66.4 & 63.6 & 65.5  \scriptsize{(\textcolor{black}{0\%})} \\

LDAM+\scriptsize{DRW}~\cite{cao2019learning}         &    62.2 & 44.1 & 27.6 & 48.8 \scriptsize{(\textcolor{black}{0\%})}  &       68.7      &        67.9         &      66.5      & 67.5 \scriptsize{(\textcolor{black}{0\%})}  \\%\vspace{2mm}\\
\hline
% \method{}{\scriptsize (CE) + DRW} \tiny{(100 epochs)}  & -& -  & - & - \scriptsize{(\textcolor{blue}{\textbf{-20.7}\%})}& - & - & - & - \scriptsize{(\textcolor{blue}{\textbf{-13.5}\%})} \\
\method{}{\scriptsize (LDAM) + DRW} \tiny{(100 epochs)}    &  63.8      &      46.5        &       29.0       &      50.8 \scriptsize{(\textcolor{blue}{\textbf{-1.7}\%})}   & 71.5 & 68.5 & 66.2 & 67.9  \scriptsize{(\textcolor{blue}{\textbf{-1.4}\%})}  \\
\method{}{\scriptsize (LDAM)+ DRW} \tiny{(200 epochs)}  &  \textbf{64.7} &  \textbf{48.2} & \textbf{31.0} & \textbf{52.2} \scriptsize{(\textcolor{blue}{\textbf{-2.9}\%})} & 71.2  & \textbf{70.6} & \textbf{69.0} & \textbf{70.0} \scriptsize{(\textcolor{blue}{\textbf{-6.3}\%})}  \\
\bottomrule
% EEL-\scriptsize{}LDAM+DRW   &   \textbf{64.3}      &      \textbf{47.9}         &     \textbf{31.4}           &    \textbf{52.0}  \\ \bottomrule       
\end{tabular}
\caption{Top-1 accuracy for DenseNet-169 trained on Imagenet LT and iNaturalist’18 datasets. The overall accuracy (All column) is decomposed into three splits corresponding to \textit{many}, \textit{medium} and \textit{few} shot settings. Numbers in parenthesis indicate the FLOPS expended by each method relative to the baseline model CE (i.e., more \textcolor{blue}{\textit{negative}} means more savings, thus better). \method{} consistently improves accuracy while expending fewer FLOPS.  
% \rd{INAT : Many=843, Med=4076, Few=3224] }
}
\label{tab:densenet_ablation}
\end{table*}

\section{Hyperparameter search}
\begin{figure}[b!]
\centering
    % \begin{minipage}[t]{0.47\textwidth}
    % \includegraphics[width=0.65\linewidth]{NEURIPS 2020/Images/acc_vs_FLOPS2.pdf}
    \includegraphics[width=0.65\linewidth]{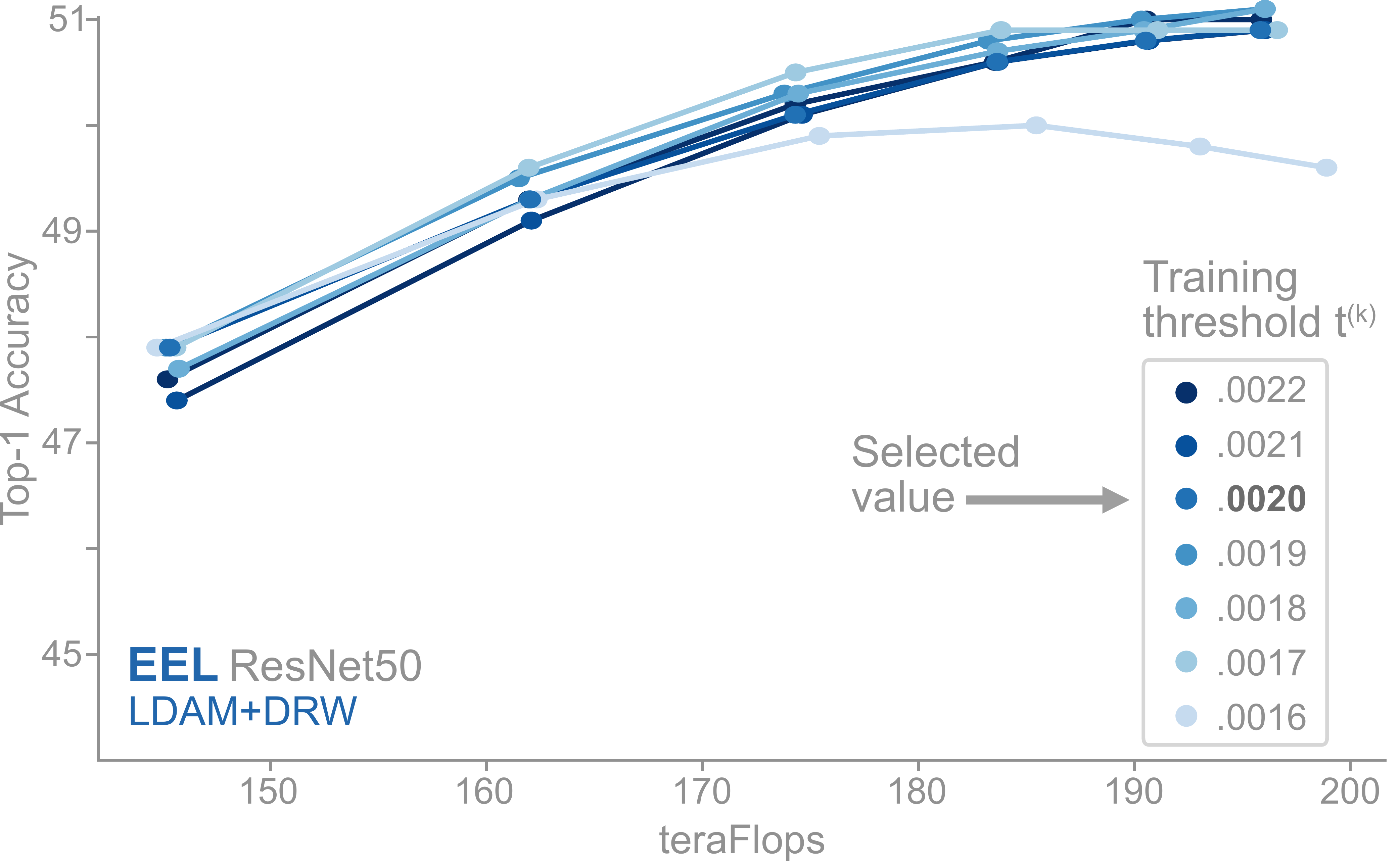}
    \caption{Ablation on training and inference thresholds $t^{(k)}, s^{(k)}$. Each point (among a total of 42) represents a Resnet-50 model trained on Imagenet LT using \method{} with different values of $(t^{(k)},s^{(k)})$. 
    Each line in the figure connects the points with the same value of $t^{(k)}$ points. 
    The tight clustering of these lines indicates that \method{} is robust to the choice of $t^{(k)}$.
    }
  \label{fig:training_ablation}
\end{figure}

We identify the best choice of the training and inference exit thresholds $t^{(k)}, s^{(k)}$ through a gridsearch. 
Figure~\ref{fig:training_ablation} summarizes the search performed for a ResNet-50 model trained on ImageNet LT dataset using \method{}{\scriptsize (LDAM)}. 
Each point in the figure corresponds to a $(t^{(k)}, s^{(k)})$ pair, showing the inference FLOPS (x-axis) and top-1 accuracy (y-axis) achieved by the corresponding model. 
In total, the Figure~\ref{fig:training_ablation} plots 42 models corresponding to seven choices of $t^{(k)} \in \{1.6,1.7,1.8,1.9,2,2.1,2.2\} \times 10^{-3}$ and six choices of $s^{(k)} \in \{1.5,1.55,1.6,1.7,1.75\} \times 10^{-3}$. 
Each line on the plot is obtained by fixing $t^{(k)}$ and varying $s^{(k)}$.
% in our search space. 
The tight clustering of lines reveals that \method{} is relatively independent to the choice of the training threshold $t^{(k)}$. 
This allows us to focus our attention to $s^{(k)}$, which can be used to generate a family of models along an efficiency-tradeoff curve.

Since \method{} is robust to $t^{(k)}$ and iterating over it is expensive (evaluating each choice involves training a new model from scratch), in practice, we fix the value of $t^{(k)}$ and iterate only over $s^{(k)}$. 
Our restricted hyperparameter search for \method{}{\scriptsize (LDAM)} is described as follows. 
We chose a fixed value of $t^{(k)} = \frac{2}{|c|}$ where $|c|$ is the number of classes in the dataset. For CIFAR-10 LT $|c|=10$, for ImageNet LT $|c|=1000$ and for iNaturalist'18 $|c|=8142$. 
The inference threshold $s^{(k)}$ is identified by searching across the set $\{1.5,1.55,1.6,1.65,1.7,1.75\} / |c|$.
Similarly, for \method{}{\scriptsize (CE)} we select a fixed value of $t^{(k)} = 0.9$ for all datasets, and the inference threshold $s^{(k)}$ is searched for across the set $\{0.5,0.55,0.6,0.65,0.7,0.75,0.8,0.85,0.9,0.95\}$.

\section{Dataset Construction}
The datasets used in this paper are constructed as follows:

\textbf{CIFAR LT datasets~\cite{cui2019class}.}
The training sets for CIFAR-10 LT, CIFAR-100 LT are sampled from the class balanced training sets of CIFAR-10 and CIFAR-100 according to the exponential distribution $n_c = n \mu^c$. Here $n_c$ refers to the remaining number of examples in class $c$, $n$ is the original number of examples per class (5000 for CIFAR-10 and 500 for CIFAR-100) and $\mu\in [0,1]$. 
We select $\mu$ such that the imbalance ratio---which is defined as the ratio between the number of examples in the largest and smallest class---is $10\times, 50\times, 100\times$.

\textbf{ImageNet LT dataset~\cite{liu2019large}.}
The training set for the ImageNet LT dataset is sampled from the original training set of ImageNet by following the pareto distribution with the $\alpha=6$. 
We follow the training split proposed by the original paper~\cite{liu2019large}.

\textbf{iNaturalist'18 dataset~\cite{van2018inaturalist}.}
This is a naturally imbalanced dataset consisting of images from 8,142 species. 
We use the same training and validation split as the original paper~\cite{van2018inaturalist}.

\section{Architecture of \method{} models}

\begin{figure}[t!]
     \centering
     \begin{subfigure}[b]{\textwidth}
         \centering
         \includegraphics[width=\textwidth]{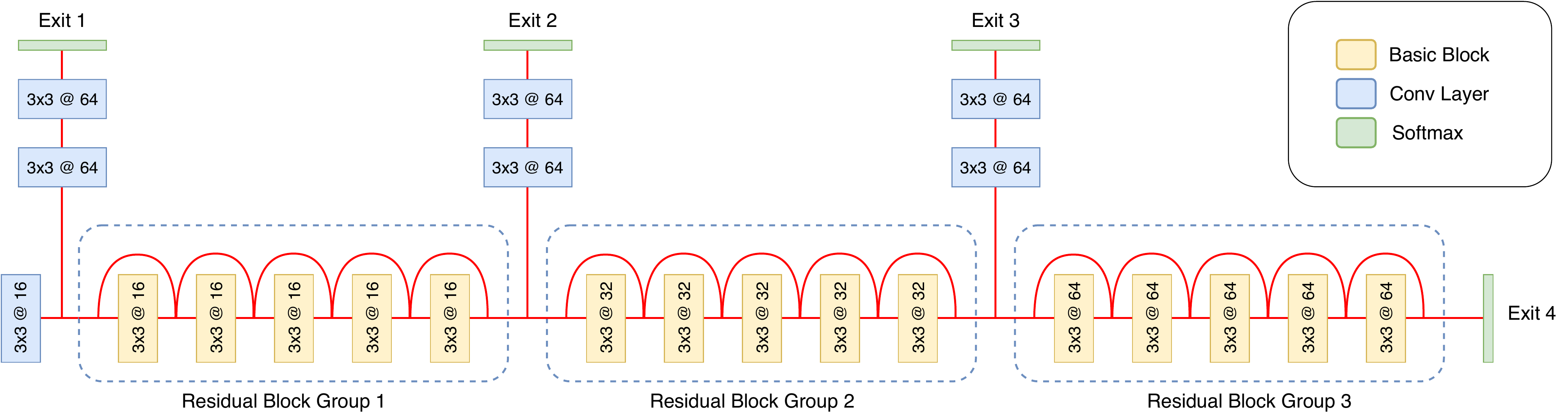}
         \caption{ResNet-32 model with early exits.  }
         \label{subfig:resnet32_elf}
         \vspace{1cm}
     \end{subfigure}
     
     \begin{subfigure}[b]{\textwidth}
         \centering
         \includegraphics[width=\textwidth]{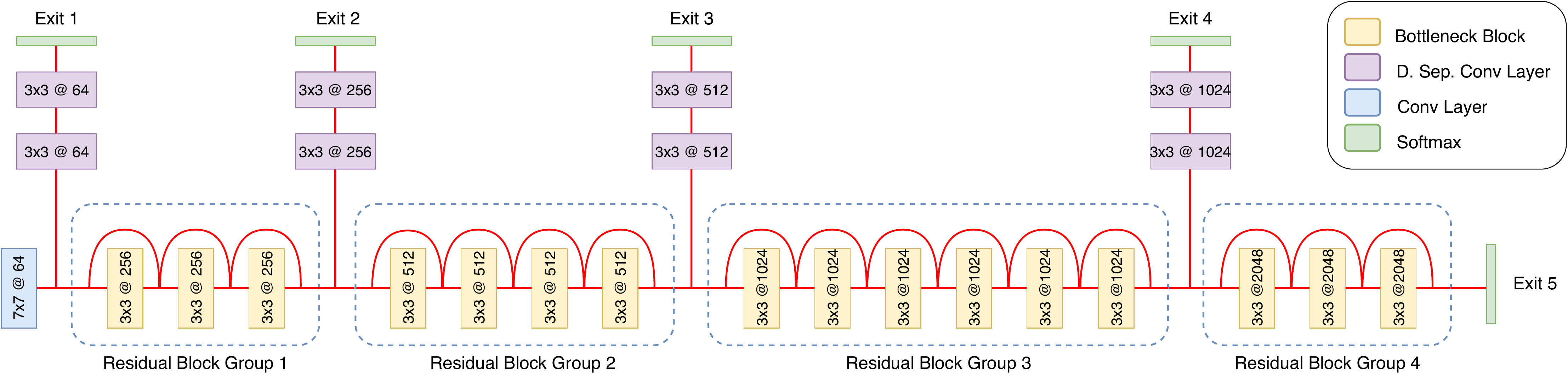}
         \caption{ResNet-50 model with early exits.}
         \label{subfig:resnet50_elf}
         \vspace{1cm}
     \end{subfigure}
     
     \begin{subfigure}[b]{\textwidth}
         \centering
         \includegraphics[width=\textwidth]{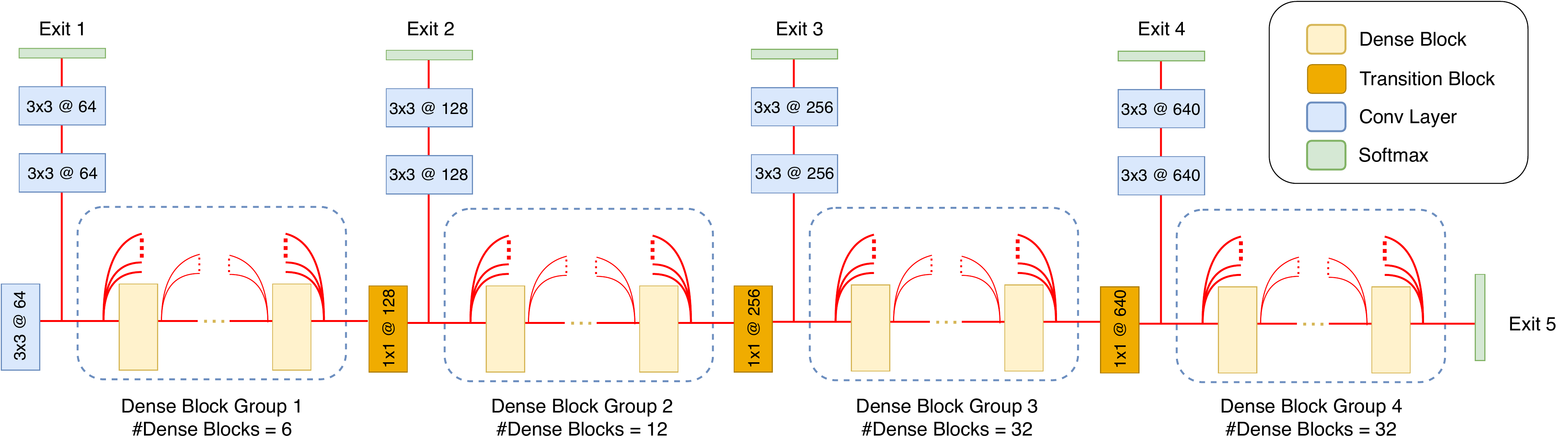}
         \caption{DenseNet-169 model with early exits.}
         \label{subfig:densenet169_elf}
     \end{subfigure}
        \caption{\method{} augments a backbone model with auxilliary exits. This figure describes the configuration of the early exits for the three models considered in this work. 
        The notation $3\times3 @ 16$ indicates that the block / layer contains 16 kernels of size $3\times3$.}
        \label{fig:elf_model_architectures}
\end{figure}

Recall that \method{} augments a backbone model with auxilliary exits. For the ResNet and DenseNet family of models, we attach an auxilliary exit before each residual/dense block. 
The augmented models are shown in Figure~\ref{fig:elf_model_architectures}, with the auxilliary exit design considerations described below.
 
\textbf{ReseNet-32}: This backbone model contains three residual block groups (see Figure~\ref{subfig:resnet32_elf}), with each group containing five standard ``basic blocks''. 
 Each auxilliary exit consists of two convolution layers with sixty-four kernels of size $3\times3$, followed by an average pooling and dense layer.

\textbf{ReseNet-50}: This backbone model contains four residual block groups (see Figure~\ref{subfig:resnet50_elf}), with the groups containing $3,4,6$ and $3$ ``bottleneck'' blocks respectively. Since the filter channels increase rapidly in this architecture (e.g. group three has 1024 channels), we use depthwise separable convolution layers at each exit which helps reduce the additional FLOPS introduced by the auxilliary exits. 
Each auxilliary exit consists of two convolution layers with $3\times3$ kernels followed by an average pooling and dense layer. The number of filters in a particular exit is the same as the number of output channels from the preceeding block.
 
\textbf{DenseNet-169}  This backbone model contains four dense block groups (see Figure~\ref{subfig:densenet169_elf}), with the groups containing $6,12,32$ and $32$ ``dense'' blocks respectively. Each auxilliary exit consists of two convolution layers with $3\times3$ kernels followed by an average pooling and dense layer. 
The number of filters in a particular exit is the same as the number of outuput channels from the preceeding block.
 
 \vspace{3cm}
 
\section{Visualizing \method's learned notion of hardness}
\begin{figure}[h!]
    \centering
    \includegraphics[width=\textwidth]{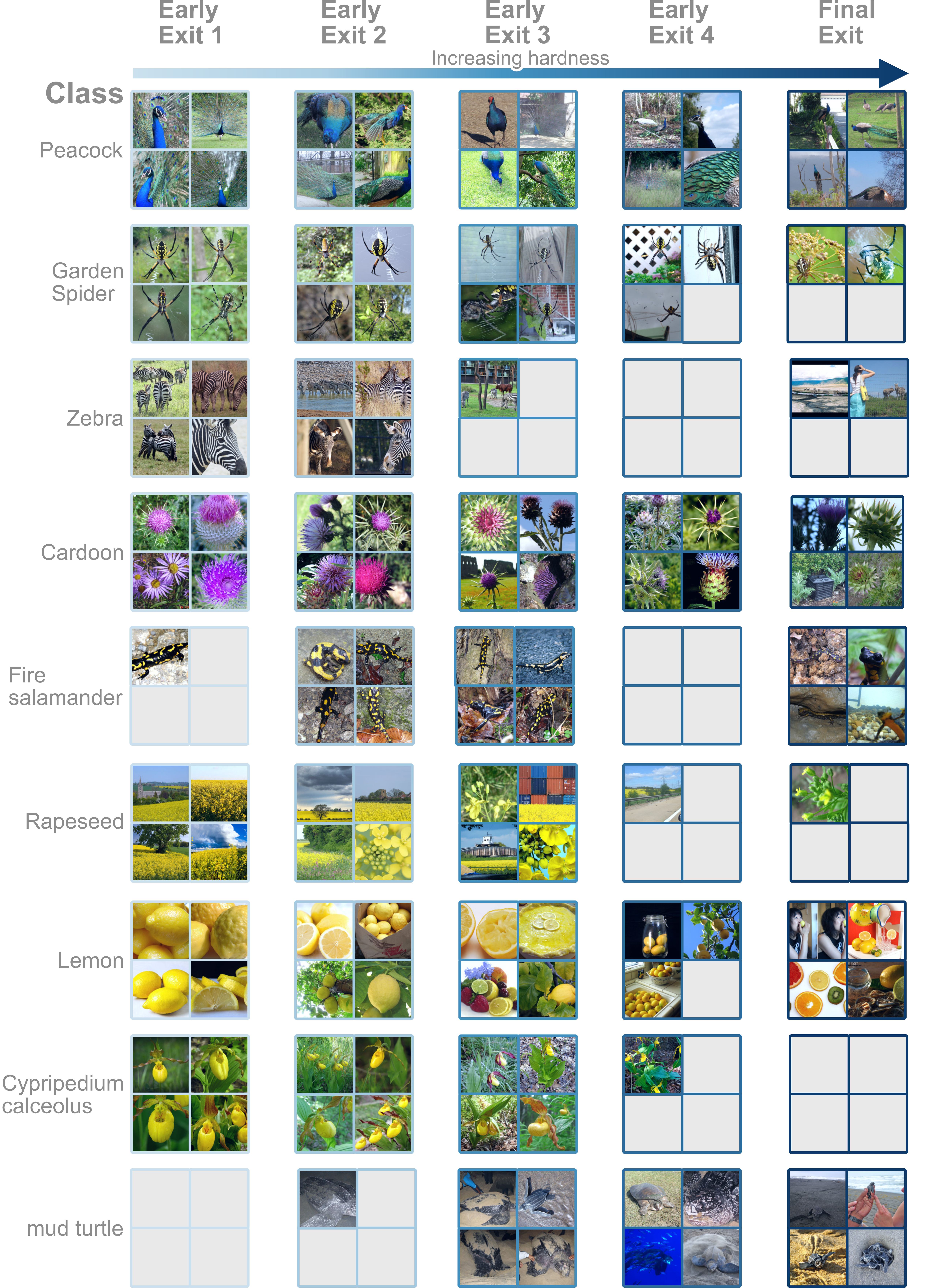}
    \caption{
    A random sample of images from the ImageNet LT dataset exiting at each auxilliary branch of a ResNet-50 \method{} model.
    As the exits increase, so does the image hardness. 
    Each class containing a gray square indicates no additional images exit from that branch.
    }
    \label{fig:collage-full}
\end{figure}

In Figure~\ref{fig:collage-full}, we provide additional example images exiting from each auxilliary exit of a ResNet-50 model. Similar to our findings in Figure~\ref{fig:image-difficulty} we can see that ``harder'' images tend to exit from the later exits, indicating that the network trained with \method{}{\scriptsize (LDAM)} indeed learns a notion of example hardness.

\end{document}